\documentclass[runningheads]{llncs}
\usepackage[T1]{fontenc}
\usepackage[utf8]{inputenc}
\usepackage{graphicx,verbatim}
\usepackage{amsmath}
\usepackage{amssymb}
\usepackage{booktabs}
\usepackage{xcolor}
\newcommand{\our}{\texttt{ColonSplat}}

\usepackage[table]{xcolor}
\begin{document}
\title{\our{}: Reconstruction of Peristaltic Motion in Colonoscopy with Dynamic Gaussian Splatting}
\titlerunning{\our{}}
%

\newcommand\blfootnote[1]{%
  \begingroup
  \renewcommand\thefootnote{}\footnote{#1}%
  \addtocounter{footnote}{-1}%
  \endgroup
}
\author{
Weronika Smolak-Dy\.{z}ewska\inst{1*} \and
Joanna Kaleta\inst{2,3*} \and
Diego Dall'Alba\inst{4,3} \and
Przemys\l{}aw Spurek\inst{1,5}
}

\authorrunning{Smolak-Dy\.{z}ewska et al.}

\institute{
Jagiellonian University, Kraków, Poland \and
Warsaw University of Technology, Poland \and
Sano Centre for Computational Medicine, Kraków, Poland \and
University of Verona, Italy, \and
IDEAS Research Institute, Warsaw, Poland \\
\email{weronika.smolak@doctoral.uj.edu.pl}
}

\authorrunning{Smolak-Dy\.{z}ewska et al.}
  
\maketitle              
\blfootnote{* Equal contribution.}

\begin{abstract}
Accurate 3D reconstruction of colonoscopy data, accounting for complex peristaltic movements, is crucial for advanced surgical navigation and retrospective diagnostics. While recent novel view synthesis and 3D reconstruction methods have demonstrated remarkable success in general endoscopic scenarios, they struggle in the highly constrained environment of the colon. Due to the limited field of view of a camera moving through an actively deforming tubular structure, existing endoscopic methods reconstruct the colon appearance only for initial camera trajectory. However, the underlying anatomy remains largely static; instead of updating Gaussians’ spatial coordinates (xyz), these methods encode deformation through either rotation, scale or opacity adjustments. In this paper, we first present a benchmark analysis of state-of-the-art dynamic endoscopic methods for realistic colonoscopic scenes, showing that they fail to model true anatomical motion. To enable rigorous evaluation of global reconstruction quality, we introduce DynamicColon, a synthetic dataset with ground-truth point clouds at every timestep. Building on these insights, we propose \our{}, a dynamic Gaussian Splatting framework that captures peristaltic-like motion while preserving global geometric consistency, achieving superior geometric fidelity on C3VDv2 and DynamicColon datasets. Project page: \url{https://wmito.github.io/ColonSplat}



\end{abstract}
\section{Introduction}
Endoluminal endoscopic procedures are central to clinical diagnosis, with colono-scopy serving as the gold standard for colorectal cancer screening. Accurate 3D reconstruction of the colon is a fundamental challenge in medical image analysis, essential for advanced endoscopic navigation, precise polyp localization, and retrospective diagnostics. The recent emergence of novel view synthesis techniques, particularly Neural Radiance Fields (NeRF) \cite{mildenhall2020nerf} and 3D Gaussian Splatting (3DGS) \cite{kerbl20233d}, has revolutionized the field of computational anatomy by enabling high-fidelity rendering of complex anatomy. Based on these foundational representations, numerous works have attempted to reconstruct colonoscopy scenes \cite{beltran2024nfl,bonilla2024gaussian,han2025endopbr,KalJoa_PRENDO_MICCAI2025,psychogyios2023realistic,wang2024endogslam}. However, a common limitation of these colon-specific methods is their reliance on a non-realistic static synthetic datasets.
By treating the colon as a rigid structure, they ignore the complex, continuous peristaltic movements that define in vivo examinations, limiting their clinical applicability.

\begin{figure}[t!]
    \centering
    \includegraphics[width=0.95\linewidth]{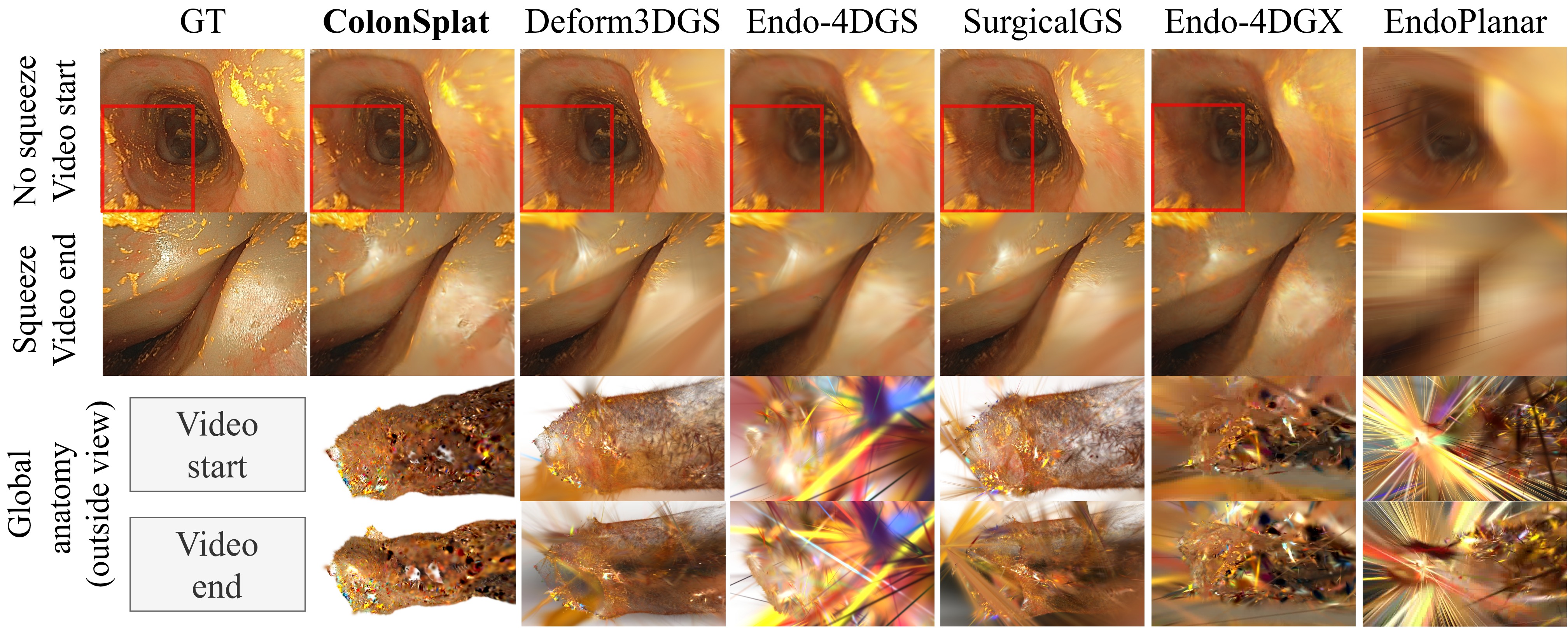}
    \caption{The C3VDv2 dataset contains realistic colonoscopy sequences with substantial non-rigid deformations over time. The top two rows show captures from the endoscopic camera and the corresponding renders produced by different methods. The bottom row illustrates the global structure of the reconstructed colon. Peristaltic-like dynamic challenge baseline approaches; however, \our{} uniquely maintains a physically plausible global structure across timesteps. \textbf{Please zoom in for details, see more examples in supplementary videos.}
    }
    \label{fig:squeeze}
\end{figure}

Parallel to these efforts, significant progress has been made in dynamic endoscopic reconstruction, successfully modeling tissue deformations based on both NeRF \cite{wang2022neural,zha2023endosurf} and 3DGS \cite{chen2025surgicalgs,huang2025endo4dgx,huang2024endo,li2024endosparse,liu2024endogaussian,paonim2025endoplanar,sunmola2025surgical}. While these dynamic methods achieve impressive results, they are primarily designed for general endoscopic or laparoscopic procedures. 
In contrast to wide-cavity scenarios with localized movement, colonoscopy navigates a constrained, narrow tube. Here, constant camera translation and non-rigid peristalsis deprive the system of persistent landmarks. Existing methods struggle with this deforming tubular topology and restricted field of view.
Due to these limitations, current models only succeed at capturing local tissue appearance consistent with views from initial camera trajectory. Global evaluation reveals geometric inconsistencies in the reconstructed colonic structure, clearly visible when looking from camera located outside of the colon (see Fig. \ref{fig:squeeze}).
Enforcing physically plausible deformations is important because: (i) overfitted Gaussians that explain motion through rotations and scale changes may lead to artifacts in novel views deviating from the training trajectory, and (ii) physically grounded reconstructions provide a more reliable foundation for integration with potential downstream physics-based simulations.

To address this crucial gap, we introduce \our{}, a novel dynamic 3DGS approach tailored specifically for the extreme conditions of colonoscopy. \our{} is designed to accurately capture the complex deformations of the tissue while explicitly maintaining the global structural integrity of the reconstructed tubular organ. Furthermore, a major obstacle in evaluating global reconstruction accuracy in colonoscopy is the lack of clinical datasets with reliable 3D ground truth of geometry. To rigorously benchmark the performance of global reconstruction, a task impossible with clinical data alone, we present a novel synthetic dataset of a dynamically moving colon with precise global trajectories and geometry. 
In summary, our main contributions are as follows:
\begin{itemize}
    \item \textbf{Benchmark Analysis:} We provide a thorough evaluation of state-of-the-art dynamic endoscopic methods in the context of colonoscopy. Our analysis shows that these approaches fail to model true anatomical motion; instead of meaningfully updating Gaussians' spatial coordinates (xyz), they encode deformation through rotation, scale and opacity, resulting in physically implausible motion reconstructions.
    \item \textbf{DynamicColon Dataset:} We introduce new dynamic dataset that provides ground-truth point clouds for every timestep. Unlike existing datasets, it enables evaluation of global geometric reconstruction and structural integrity in the presence of tissue deformations.
    \item \textbf{\our{} Method:} We propose novel dynamic reconstruction framework that successfully captures complex peristaltic-like motion while preserving global geometric consistency, achieving superior geometric fidelity on both the highly realistic C3VDv2 \cite{c3vd2} and DynamicColon datasets.
\end{itemize}

\section{Related Work}

\textbf{Static Novel View Synthesis in Colonoscopy.}
The introduction of NeRF \cite{mildenhall2020nerf} and 3DGS \cite{kerbl20233d} has significantly advanced the capabilities of novel view synthesis in medical imaging. In the specific context of colonoscopy, several methods have been developed to reconstruct the complex geometry and texture of the colon interior. NeRF-based approaches, such as REIM-NERF \cite{psychogyios2023realistic} and NFL-BA \cite{beltran2024nfl}, utilize implicit neural representations to model the tissue. More recently, 3DGS has been adapted for colonoscopy to leverage its explicit representation and real-time rendering capabilities, as seen in Gaussian Pancakes \cite{bonilla2024gaussian}, EndoGSLAM \cite{wang2024endogslam}, PR-ENDO \cite{KalJoa_PRENDO_MICCAI2025}, and EndoPBR \cite{han2025endopbr}. While these methods achieve high fidelity reconstructions, they fundamentally assume an unnaturally static environment. By modeling the colon as a rigid structure, they cannot account for the continuous peristaltic motion observed during in vivo procedures, thereby limiting their applicability in realistic dynamic scenarios.

\textbf{Dynamic Reconstruction in General Endoscopy.} To address tissue deformation during surgery, various dynamic reconstruction techniques have been proposed, primarily focusing on general endoscopic and laparoscopic procedures. Early dynamic methods relied on NeRF architectures \cite{wang2022neural,zha2023endosurf} to model non-rigid deformations over time. However, the implicit nature of these models often results in computationally expensive training and rendering, making it difficult to explicitly track tissue movement in real-time clinical applications.

A multitude of recent works, including EndoPlanar \cite{paonim2025endoplanar}, Endo-4DGS \cite{huang2024endo}, ENDO-4DGX \cite{huang2025endo4dgx}, SurgicalGS \cite{chen2025surgicalgs}, SGS \cite{sunmola2025surgical}, EndoGaussian \cite{liu2024endogaussian}, and EndoSparse \cite{li2024endosparse}, have demonstrated exceptional performance in modeling surgical scenes using dynamic 3DGS-based frameworks. While these methods effectively capture dynamic tool-tissue interactions, they are mainly designed for cavity-like environments where the region of interest remains consistently visible. In such laparoscopic settings, deformations typically unfold within a broad field of view that provides sufficient geometric overlap for spatial consistency. 

\textbf{Addressing the Challenges of Dynamic Colonoscopy.} Despite the success of dynamic 3DGS in laparoscopy, directly applying these state-of-the-art methods to dynamic colonoscopy reveals significant limitations. Colonoscopy presents a unique set of challenges characterized by a highly constrained tubular topology, a limited field of view, and global peristaltic motion. As the camera continuously translates forward, previously observed regions quickly leave the field of view. 

\begin{figure}[t!]
    \centering
    \includegraphics[width=\linewidth]{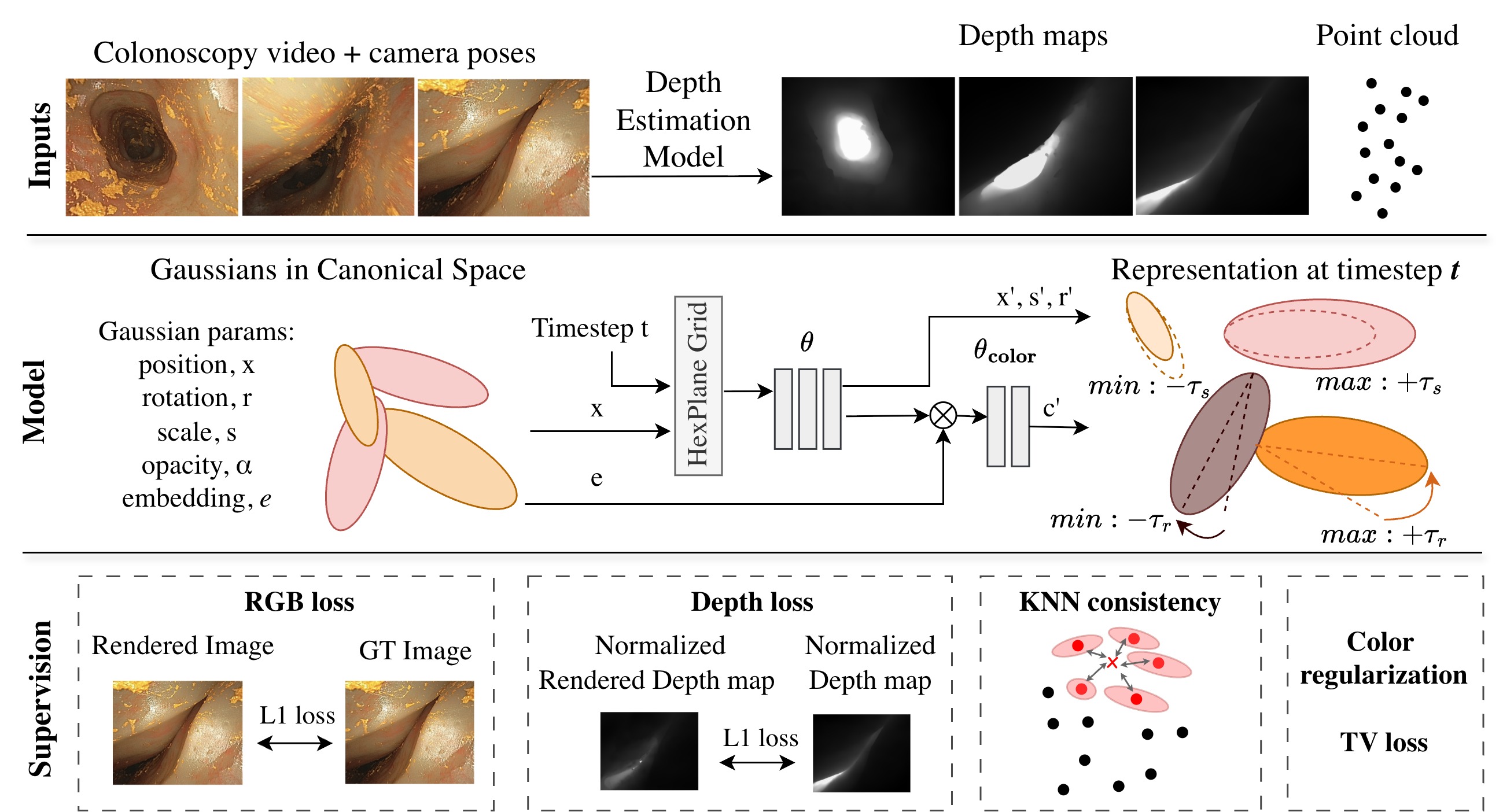}
    \caption{\our{} reconstructs dynamic 3D anatomy from colonoscopy video using estimated depth. A deformation model updates canonical Gaussians parameters at each time step for consistent dynamic reconstruction. Training uses RGB and depth supervision, with KNN and color regularization, to ensure accurate, artifact-free results.}
    \label{fig}
\end{figure}

Consequently, when existing dynamic endoscopic methods are deployed in this environment, they successfully reconstruct the tissue appearance on a purely local scale but fail to maintain global geometric consistency. The lack of persistent global landmarks, combined with continuous non-rigid deformation and forward camera translation, leads to severe geometric inconsistencies and profound structural artifacts in these models, especially when viewed globally. Our proposed method, \our{}, focuses on reconstructing movement of a colon as global changes to geometry rather than local overfitting to initial camera trajectory.

\section{Method}
This section presents the components of our method, summarized in Fig.~\ref{fig}.  

\textbf{Gaussian Splatting Representation.} We represent the scene as a collection of 3D anisotropic Gaussians rendered using differentiable rasterization. Each Gaussian in canonical space is parameterized by a mean position $\mathbf{x} \in \mathbb{R}^3$, scale $\mathbf{s} \in \mathbb{R}^3$, rotation $\mathbf{r} \in \mathbb{R}^4$, opacity $\alpha \in \mathbb{R}^1$, and color $\mathbf{c} \in \mathbb{R}^3$. Additionally, each Gaussian has a learnable embedding vector $\mathbf{e}_i \in \mathbb{R}^d$ used exclusively for color prediction. Given calibrated RGB images and camera poses, we use a monocular depth estimator (ColonCrafter \cite{hardy2025coloncrafter} or AnyDepth \cite{ren2026anydepth}) to predict depth maps, from which we initialize the Gaussian point cloud. To render novel views Gaussians are projected to the image plane and alpha-composited in depth order.

\textbf{Dynamic Deformation Modeling.} Colon motion is modeled with a multilayer perceptron $\theta$ that predicts time-dependent geometric updates. Given timestep $t \in [0,1]$ and canonical Gaussian parameters, we predict geometric offsets:
\begin{equation}
    \textstyle 
    {
        (\Delta \mathbf{x}, \Delta \mathbf{s}, \Delta \mathbf{r})
        =
        \theta\bigl(\mathrm{G}(t), \mathrm{G}(\mathbf{x})\bigr).
    }
\end{equation}
$G$ denotes encoding with HexPlane Grid, following \cite{Cao2023HexPlane,Wu_2024_CVPR}. The deformed geometric parameters are obtained additively:
\begin{equation}
    \textstyle 
    {
        \mathbf{x}' = \mathbf{x} + \Delta \mathbf{x}, \quad
        \mathbf{s}' = \mathbf{s} + \mathrm{clip}(\Delta \mathbf{s}, -\tau_s, \tau_s), \quad
        \mathbf{r}' = \mathbf{r} + \mathrm{clip}(\Delta \mathbf{r}, -\tau_r, \tau_r).
    }
\end{equation}
Clipping thresholds $\tau_s$ and $\tau_r$ restrict the magnitude of scale and rotation updates, preventing these parameters from compensating for insufficient spatial deformation. Opacity $\alpha$ is fixed over time to prevent trivially explaining motion by hiding and revealing Gaussians. Additionally, updated Gaussian scales $s'$ are limited to 5\% of the scene extent to prevent oversized primitives.

\textbf{Color Modeling.} Endoscopic imaging exhibits strong appearance variations due to the collocated camera and light source, as well as fluid-related effects in realistic colon environments. Moreover, local tissue characteristics (e.g., fluids or polyps) may exhibit distinct temporal reflectance behavior. To account for these illumination and tissue-specific effects, color is modeled separately using a dedicated branch of the network, denoted as $\theta_{\mathrm{color}}$, which leverages the per-Gaussian embedding $\mathbf{e}_i$ to capture localized temporal appearance variations. Color updates are modeled multiplicatively to capture light-surface interactions:

\begin{equation}
    \textstyle 
    {
        \Delta \mathbf{c}
        =
        \theta_{\mathrm{color}}
        \bigl(\mathrm{G}(t), \mathrm{G}(\mathbf{x}), \mathbf{e}_i \bigr),
        \quad
        \mathbf{c}' =
        \mathbf{c} \odot (1 + \Delta \mathbf{c}).
    }
\end{equation}

Following \cite{chen2025surgicalgs}, we optimize $\ell_1$ RGB reconstruction loss $\mathcal{L}_{\mathrm{rgb}}$ between the rendered and GT images, and apply a total variation loss $\mathcal{L}_{\mathrm{TV}}$ on the renders to promote spatial smoothness and reduce high-frequency artifacts.

\textbf{K-Nearest Neighbor Deformation Consistency. } To encourage locally coherent non-rigid motion, for each Gaussian $i$ we identify its $K$ nearest neighbors $\mathcal{N}_i$ in canonical space and penalize inconsistencies in their deformed positions:
\begin{equation}
    \textstyle 
    {
        \mathcal{L}_{\mathrm{knn}} =
        \frac{1}{N} \sum_i
        \left\|
        \mathbf{x}'_i(t) - \frac{1}{|\mathcal{N}_i|}
        \sum_{j \in \mathcal{N}_i}
        \mathbf{x}'_j(t)
        \right\|_2^2.
    }
\end{equation}  

\textbf{Color regularization.} To prevent motion from being spuriously explained through color variation, we penalize predicted color offsets ($L_{co}$). 
Additionally, we slighlty penalize color variance across all Gaussians to avoid implausible color artifacts when only partial geometry is observed ($L_{cv}$).
\begin{equation}
    \textstyle 
    {
        \mathcal{L}_{\mathrm{co}}
        =
        \frac{1}{N}\sum_i \left\|\Delta \mathbf{c}_i\right\|_2^2,
        \qquad
        \mathcal{L}_{\mathrm{cv}}
        =
        \frac{1}{N}\sum_i
        \left\|
        \mathbf{c'}_i - \frac{1}{N}\sum_j \mathbf{c'}_j
        \right\|_2^2.
    }   
\end{equation}

\textbf{Depth regularization.} Geometry is supervised using monocular depth estimates by applying an $\ell_1$ loss between normalized supervision depth and rendered expected depth:
\begin{equation}
    \textstyle 
    {
        \mathcal{L}_{\mathrm{depth}} =
        \left\|
        D_{\mathrm{rendered}}^{\mathrm{norm}}
        -
        D_{\mathrm{sup}}^{\mathrm{norm}}
        \right\|_1.
    }
\end{equation}  

\textbf{Training Objective.} The objective combines reconstruction and regularizers:
\begin{equation}
    \textstyle 
    {
        \mathcal{L} =
        \mathcal{L}_{\mathrm{rgb}}
        + \lambda_{\mathrm{TV}}
        \mathcal{L}_{\mathrm{TV}} 
        + \lambda_{\mathrm{knn}} \mathcal{L}_{\mathrm{knn}}
        + \lambda_{\mathrm{depth}} \mathcal{L}_{\mathrm{depth}}
        + \lambda_{\mathrm{co}} \mathcal{L}_{\mathrm{co}}
        +
        \lambda_{\mathrm{cv}} \mathcal{L}_{\mathrm{cv}}.
    }
\end{equation}

\section{Experiments}
We provide data, code and all implementation details in our GitHub repository\footnote{\url{https://anonymous.4open.science/r/colonsplat-710E/README.md}}.  

\textbf{Datasets.}
Our evaluation utilizes the C3VDv2 dataset \cite{c3vd2} for its high fidelity. It introduces challenging visual artifacts like blood and mucus, while its simulated peristaltic-like motion creates the significant global geometric deformations similar to live colonoscopies.
These characteristics make reconstruction significantly more challenging than previously studied endoscopic benchmarks. We used 9 sequences from C3VDv2 with deformation for which ground truth camera poses were available. 
Following standard practice, we use every 8th frame for testing and the remaining frames for training.

Additionally, we introduce \textbf{DynamicColon} dataset created from C3VDv2 meshes, custom textures and cage based colon deformations. It consists of three scenes with train and test camera trajectories, camera views from outside, depth maps and point clouds for every deformation step which enables precise evaluation of geometric and deformation fidelity. Dataset is provided on our repository.

\begin{table}[t]
\centering
\fontsize{8pt}{9.6pt}\selectfont
\setlength{\tabcolsep}{0.75pt}
\caption{Quantitative comparison and ablation study on C3VDv2 and DynamicColon. Compared to baselines \our{} achieve superior geometry fidelity while offering strong reconstruction results. Ablation is shown in three bottom rows. *Lack of $\Delta c$ modeling caused major Gaussian drift and degraded CH and HD95 in one scene.}
\label{tab:quantitative}
\begin{tabular}{lccc|cccccc}
\toprule
& \multicolumn{3}{c|}{\textbf{C3VDv2}} 
& \multicolumn{6}{c}{\textbf{DynamicColon}} \\
\cmidrule(lr){2-4} \cmidrule(lr){5-10}
Method &
PSNR $\uparrow$ & SSIM $\uparrow$ & LPIPS $\downarrow$ &
PSNR $\uparrow$ & SSIM $\uparrow$ & LPIPS $\downarrow$ &
CH $\downarrow$ & HD95 $\downarrow$ & MSE$_{D}$ $\downarrow$  \\
\midrule
EndoPlanar & 27.229 & 0.738  & 0.499 & 28.195 & 0.902 & 0.332  & 0.497 & \cellcolor{yellow!15}{1.055} & 0.2999\\
SurgicalGS & \cellcolor{orange!25}{27.624} & \cellcolor{red!25}{0.766} & \cellcolor{red!25}{0.350} & \cellcolor{orange!25}{31.635} & \cellcolor{yellow!15}{0.942} & \cellcolor{red!25}{0.075} & 0.384 & 1.136  & \cellcolor{yellow!15}{0.0079} \\
Deform3DGS & 27.419 & \cellcolor{yellow!15}{0.745} & \cellcolor{orange!25}{0.395} & \cellcolor{orange!25}{31.635} & \cellcolor{yellow!15}{0.942} & \cellcolor{orange!25}{0.109} & \cellcolor{yellow!15}{0.383} & 1.143 & \cellcolor{orange!25}{0.0077} \\
Endo4DGS & 27.491 & 0.728 & 0.582 & \cellcolor{yellow!15}{30.548} & \cellcolor{orange!25}{0.945} & 0.155 & \cellcolor{orange!25}{0.372} & \cellcolor{orange!25}{1.017} &  0.0080 \\
Endo4DGX & \cellcolor{yellow!15}{24.525} &	0.707 & 0.521 & 29.722 &	0.921 & 0.322 & 0.433 & 1.133 & 0.0169 \\
\midrule
\our{} & \cellcolor{red!25}{28.281} & \cellcolor{orange!25}{0.750} & \cellcolor{yellow!15}{0.444} & \cellcolor{red!25}{33.633} & \cellcolor{red!25}{0.955} & \cellcolor{yellow!15}{0.137} & \cellcolor{red!25}{0.162} & \cellcolor{red!25}{0.730} & \cellcolor{red!25}{0.0060}  \\
\midrule
w/o Constraints & \cellcolor{gray!15}{27.624} & \cellcolor{gray!15}{0.736} & \cellcolor{gray!15}{0.523} & \cellcolor{gray!15}{34.536} & \cellcolor{gray!15}{0.959} & \cellcolor{gray!15}{0.126} & \cellcolor{gray!15}{0.234} & \cellcolor{gray!15}{0.854}  & \cellcolor{gray!15}{0.0055} \\
w/o $L_{KNN}$ & \cellcolor{gray!15}{28.452} & \cellcolor{gray!15}{0.752} & \cellcolor{gray!15}{0.438} & \cellcolor{gray!15}{30.548} & \cellcolor{gray!15}{0.945} & \cellcolor{gray!15}{0.145} & \cellcolor{gray!15}{0.981} & \cellcolor{gray!15}{1.033} & \cellcolor{gray!15}{0.0060} \\
w/o $\Delta c$ modelling & \cellcolor{gray!15}{27.579} & \cellcolor{gray!15}{0.733} & \cellcolor{gray!15}{0.496} & \cellcolor{gray!15}{25.857} & \cellcolor{gray!15}{0.843} & \cellcolor{gray!15}{0.342} & \cellcolor{gray!15}{11.207*} & \cellcolor{gray!15}{3.544*} & \cellcolor{gray!15}{0.0305} \\
\bottomrule
\end{tabular}
\end{table}

\begin{figure}[t!]
    \centering
    \includegraphics[width=\linewidth]{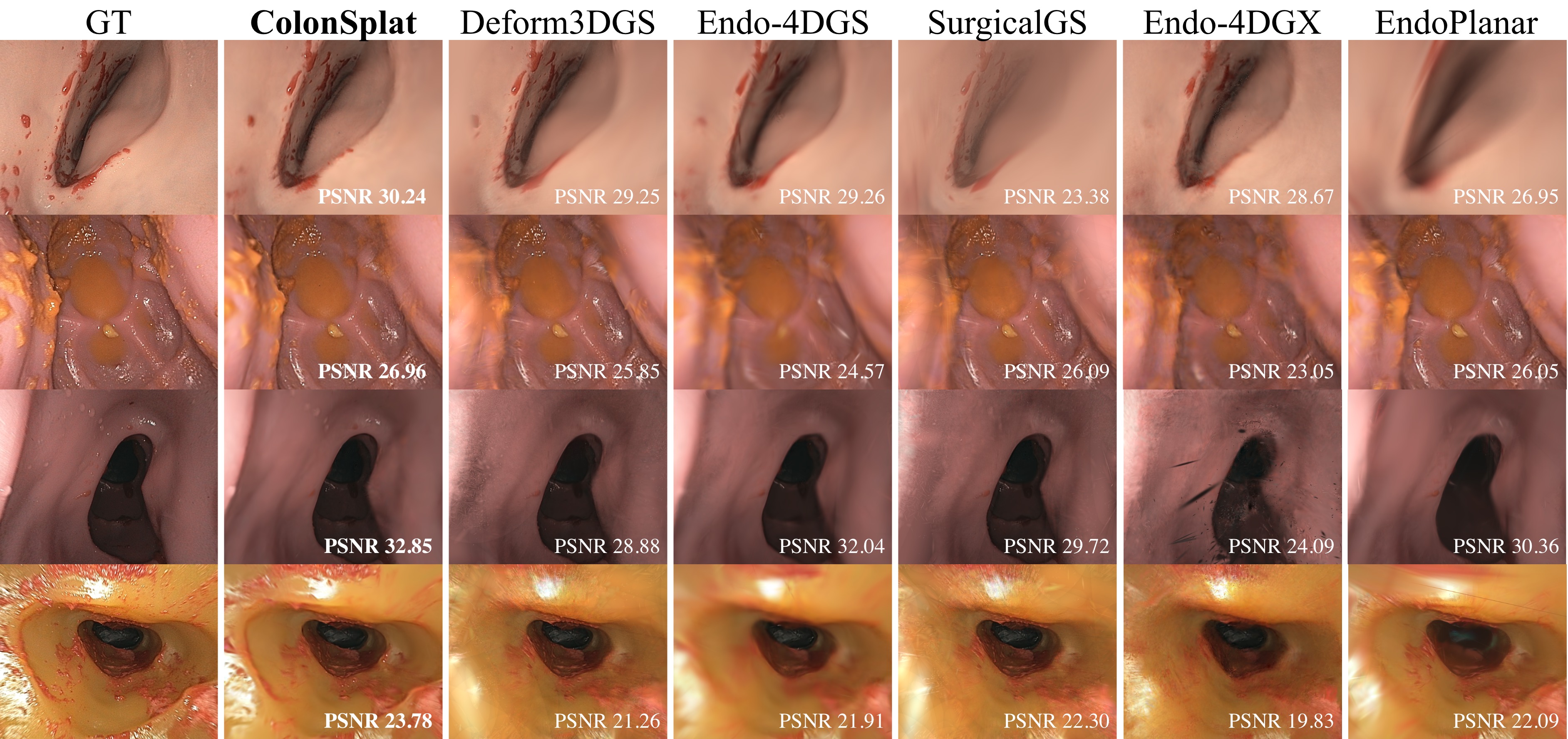}
    \caption{Qualitative comparison for C3VDv2 dataset. \textbf{Please zoom in for details.} \our{} achieves superior reconstruction quality compared to the baselines.}
    \label{fig:qualitative_comparison}
\end{figure}

\begin{figure}[t]
    \centering
    \includegraphics[width=\linewidth]{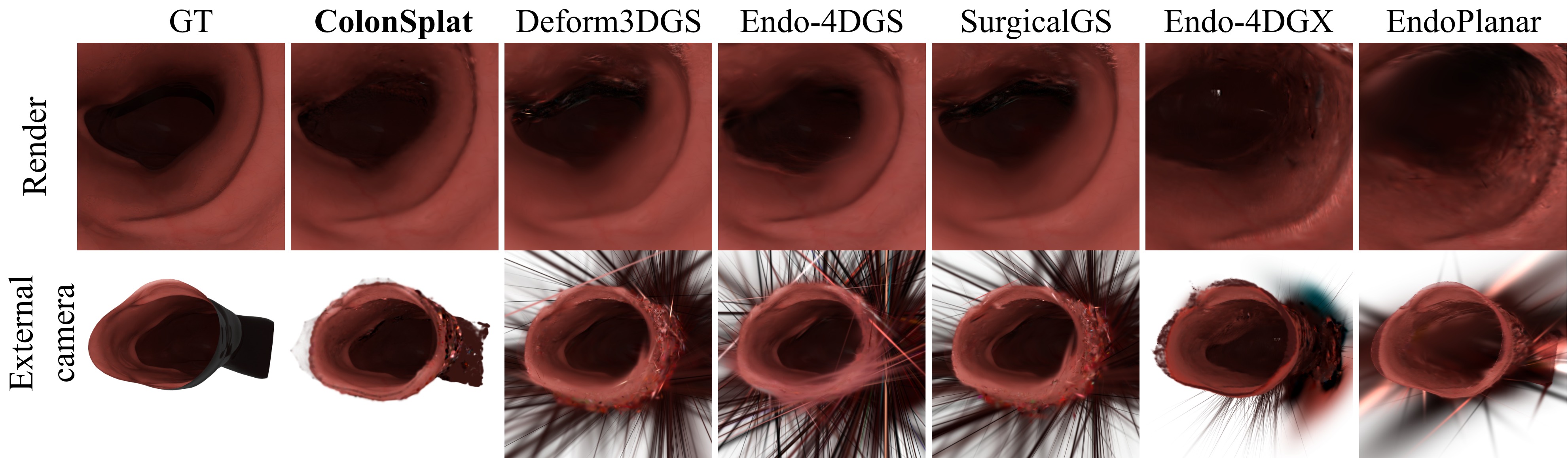}
    \caption{DynamicColon. The top row shows renders from test trajectory. The bottom row presents views from cameras positioned outside the colon. Competing methods produce significant artifacts that mimic deformations. \our{} accurately captures the 3D structure of the colon without such artifacts, even when viewed from outside.}
    \label{fig:qualitative_custom}
\end{figure}

\begin{figure}[h!]
    \centering
    \includegraphics[width=\linewidth]{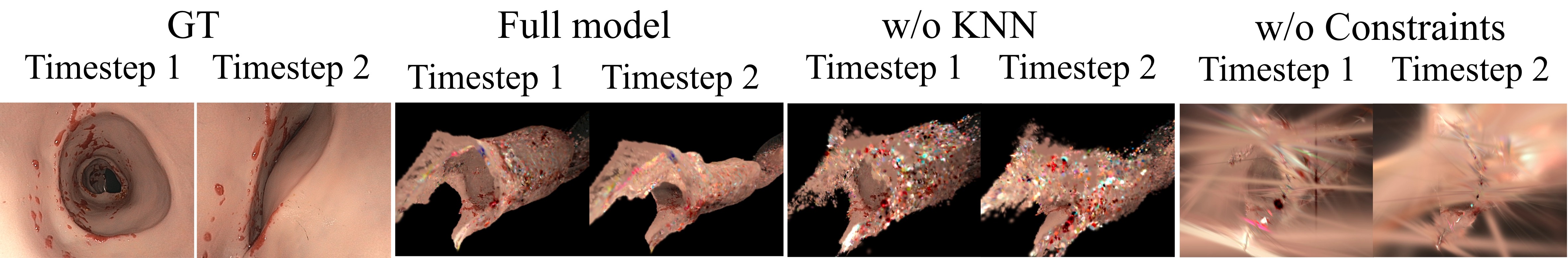}
    \caption{Ablation study on C3VDv2 scene. Our proposed components significantly enhance realistic dynamic colon deformation. KNN regularization provides smooth tissue-like Gaussian surface across timesteps while our proposed constraints eliminate geometric artifacts. \textbf{Please zoom in for details. }}
    \label{fig:ablation}
\end{figure}


\textbf{Baselines.}
We compare ColonSplat with several recent dynamic endoscopic reconstruction methods: Endo4DGS \cite{huang2024endo}, Deform3DGS \cite{Yan_Deform3DGS_MICCAI2024}, SurgicalGS \cite{chen2025surgicalgs}, EndoPlanar \cite{paonim2025endoplanar}, Endo4DGX \cite{huang2025endo4dgx}. As these methods handle depth differently - some relying on normalized depth, others 
 on metric depth - their original depth losses are not always directly compatible with ColonCrafter depth maps. This in practice led to convergence issues. To ensure a fair comparison, we apply the same L1 depth loss on normalized depth maps across all methods. 
We also tune depth loss weights and training iterations for each baseline to obtain the most satisfactory renderings and consistent geometry for each method to ensure fair comparison. 

\textbf{Quantitative Comparison.}
Tab.~\ref{tab:quantitative} reports quantitative results across all baselines and \our{}. For C3VDv2, we report standard image reconstruction metrics computed on held-out test frames: PSNR, SSIM, and LPIPS.   

\our{} achieves strong reconstruction quality compared to baselines. For the DynamicColon with available ground-truth geometry, we additionally report Chamfer Distance (CD) and 95th percentile Hausdorff Distance (HD95) on a point cloud sampled from Gaussian representation at each test timestep, and MSE on normalized render depths $\text{MSE}_{D}$. These metrics enable direct evaluation of deformation fidelity. \our{} consistently improves geometric consistency under strong peristaltic-like deformation. 

\textbf{Qualitative Results.}
We present qualitative comparisons in Fig.~\ref{fig:qualitative_comparison} and \ref{fig:qualitative_custom}.
\our{} produces more stable large-scale geometry and fewer view-dependent artifacts compared to prior methods, particularly in regions affected by strong tissue deformation. \textbf{We strongly encourage to view our supplement.}  

\textbf{Ablation Study.} To evaluate the contribution of individual components, we perform ablations under three settings:
(i) \emph{No Constraints}, where all Gaussian parameters are freely optimized without structural regularization;
(ii) \emph{No KNN Consistency}; and
(iii) \emph{No Color Modeling ($\Delta \mathbf{c})$}. Quantitative results are reported in Tab.~\ref{tab:quantitative} and qualitative results are shown in Fig.~\ref{fig:ablation}.

\section{Conclusions}
\our{} introduces a dynamic Gaussian Splatting framework tailored for colonoscopy that captures complex motion while preserving global anatomical consistency. By explicitly constraining deformation and incorporating additional supervision, it mitigates the structural artifacts observed in prior dynamic baselines. Extensive analysis of state-of-the-art baselines on C3VDv2 and the proposed DynamicColon dataset demonstrates superior geometric fidelity, strong reconstruction quality and provides evaluation framework for future works.  

\textbf{Limitations:} Despite significant improvements, motion reconstruction can remain inaccurate when deformations are very poorly observed in the video.

\textbf{Acknowledgements.} Joanna Kaleta is supported by National
Science Centre, Poland (grant no. 2022/47/O/ST6/01407). This paper received funding from the European Union’s Horizon 2020 research and innovation programme under grant agreement No 857533. The research is supported by Sano project carried out within the International Research Agendas programme of the Foundation for Polish Science, co-financed by the European Union under the European Regional Development Fund. The research was created within the project of the Minister of Science and Higher Education ”Support for the activity of Centers of Excellence established in Poland under Horizon 2020” on the basis of the contract number MEiN/2023/DIR/3796. The work of W. Smolak-Dy\.{z}ewska and P. Spurek was supported by the project \textit{Effective Rendering of 3D Objects Using Gaussian Splatting in an Augmented Reality Environment} (FENG.02.02-IP.05-0114/23), carried out under the First Team programme of the Foundation for Polish Science and co-financed by the European Union through the European Funds for Smart Economy 2021–2027 (FENG). 
%

\bibliographystyle{splncs04}

\end{document}